\documentclass{article}
\usepackage{spconf,amsmath,graphicx}
\usepackage{amsfonts}
\usepackage{graphicx}
\usepackage{booktabs}
\usepackage[labelfont=small,textfont=small]{caption,subfig}
\usepackage{tikz}
\usepackage{pstricks}
\usetikzlibrary{spy}
\usepackage{pgfplots}
\pgfplotsset{compat=newest}
\pgfplotsset{plot coordinates/math parser=false}
\newlength \figureheight
\newlength \figurewidth
\definecolor{bblue}{HTML}{4F81BD}
\definecolor{rred}{HTML}{C0504D}
\definecolor{ggreen}{HTML}{00AA00}

\def\argmin{\mathop{\mathrm{argmin}}}

\pgfplotsset{
compat=1.11,
legend image code/.code={
\draw[mark repeat=2,mark phase=2]
plot coordinates {
(0cm,0cm)
(0.15cm,0cm) 
(0.3cm,0cm) 
};%
}
}

\usepackage{hyperref}

\title{Learning from a Handful Volumes: MRI Resolution Enhancement with Volumetric Super-Resolution Forests}

\name{Aline Sindel$^1$Katharina Breininger$^1$Johannes K\"a{\ss}er$^2$Andreas Hess$^2$Andreas Maier$^{1,3}$Thomas K\"ohler$^{1,3,4}$}
\address{$^1$Pattern Recognition Lab, FAU Erlangen-N\"urnberg, Germany\\
\fontdimen2\font=0.15em $^2$Institute for Experimental \& Clinical Pharmacology \& Toxicology, FAU Erlangen-N\"urnberg, Germany \\
\hspace{-0.8em}\fontdimen2\font=0.15em $^3$Erlangen Graduate School in Advanced Optical Technologies (SAOT), FAU Erlangen-N\"urnberg, Germany \\
$^4$e.solutions GmbH, Erlangen, Germany\\[0.2em]
\url{https://github.com/asindel/VSRF}
}

\begin{document}
\ninept
\maketitle
\begin{abstract}
Magnetic resonance imaging (MRI) enables 3-D imaging of anatomical structures. However, the acquisition of MR volumes with high spatial resolution leads to long scan times. To this end, we propose volumetric super-resolution forests (VSRF) to enhance MRI resolution retrospectively. Our method learns a locally linear mapping between low-resolution and high-resolution volumetric image patches by employing random forest regression. We customize features suitable for volumetric MRI to train the random forest and propose a median tree ensemble for robust regression. VSRF outperforms state-of-the-art example-based super-resolution in term of image quality and efficiency for model training and inference in different MRI datasets. It is also superior to unsupervised methods with just a handful or even a single volume to assemble training data. 
\end{abstract}
\section{Introduction}
\label{sec:intro}
High-resolution magnetic resonance imaging (MRI) allows the visualization of delicate anatomical structures in-vivo, which is crucial to support early detection of pathologies and to enable an accurate prediction of their size and composition.
However, high-resolution MRI for this task requires long acquisition times, which can lead to stress and discomfort for the imaged subject. To reduce acquisition times while retaining high spatial resolution, resolution enhancement can be applied during data acquisition, e.g. using zero-filling, or retrospectively by means of super-resolution (SR)~\cite{Reeth2012}. 

SR estimates high-resolution (HR) images from single or sets of low-resolution (LR) images. \textit{Multi image} SR \cite{Koehler2016,Bercea2016, Koehler2017} can effectively enhance the spatial resolution but multiple acquisitions increase scan times. \textit{Single image} (SISR) methods are promising alternatives. \textit{Reconstruction-based} SISR methods \cite{Dai2007,Bareja2012} are based on a regularized optimization problem to enforce the downsampled version of the predicted HR result to be close to the LR image. \textit{Example-based} methods estimate a HR image from a single LR image based on pairs of LR/HR examples of an external database. In this area, 
dictionary-based approaches \cite{Yang2010, Zeyde2012, Timofte2013, Timofte2015} build on sparse representations of image patches. Regression-based approaches are more effective as they avoid time-consuming sparse coding. These include tree-based regression \cite{Schulter2015} on a patch level as well as deep learning \cite{Dong2014,Kim2016,Lim2017} to infer end-to-end mappings. Another class of SISR, \textit{self} SR \cite{Glasner2009, Freedman2010}, estimates the HR image by internally learning from patches of different scales without external databases.

\begin{figure}[!t]
	\centering
	\mbox{
	\hspace{-1.3em}
	\subfloat[LR image]{
		\begin{tikzpicture}[spy using outlines={rectangle,red,magnification=2.0,height=1.8cm, width=1.8cm, connect spies, every spy on node/.append style={thick}}] 
			\node {\pgfimage[width=0.45\linewidth]{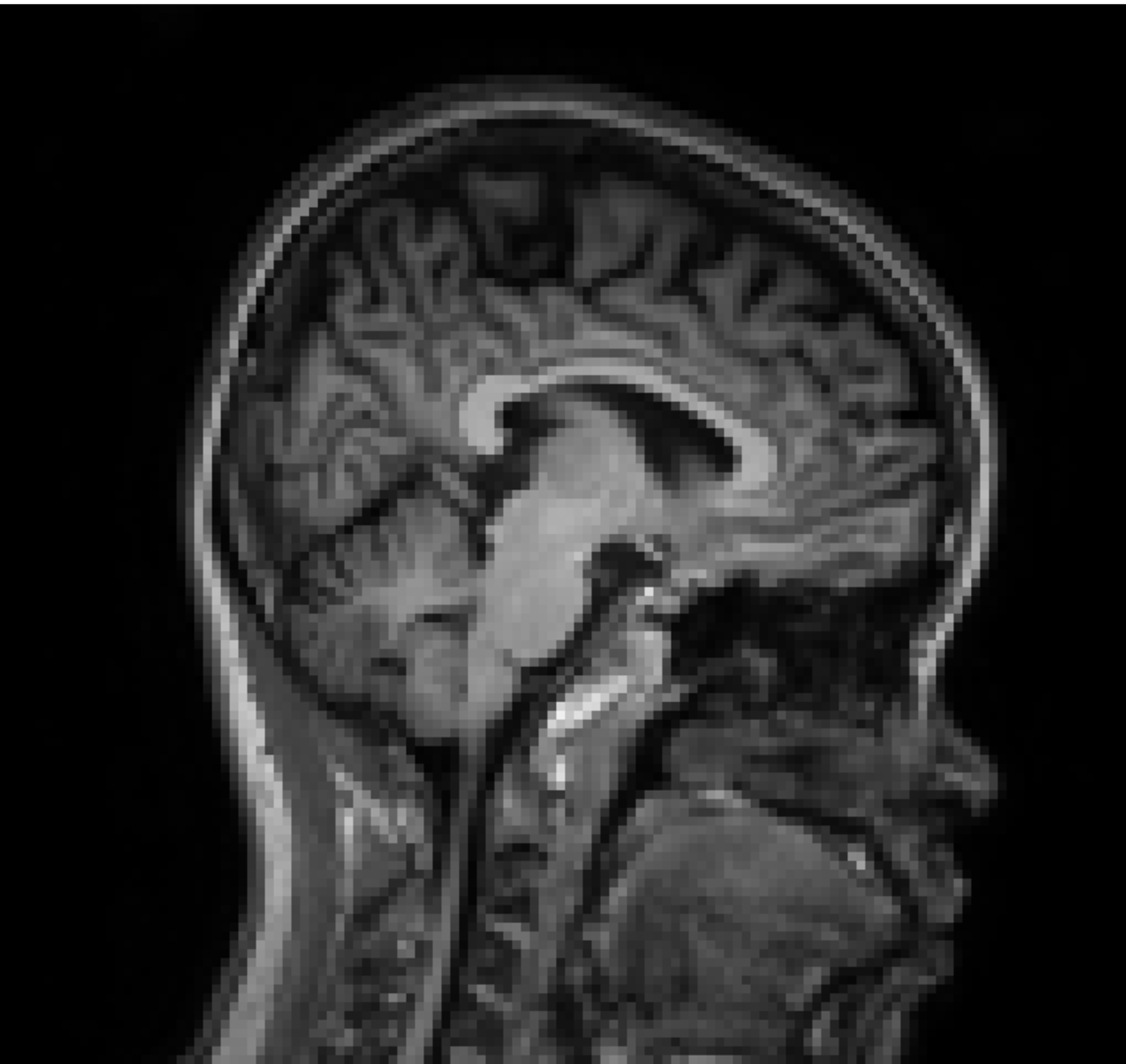}}; 
      		\spy on (-0.70, 0) in node [left] at (1.915, -0.895); 
    	\end{tikzpicture}
	}\hspace{0.1em}
	\subfloat[VSRF (ours)]{
		\begin{tikzpicture}[spy using outlines={rectangle,red,magnification=2.0,height=1.8cm, width=1.8cm, connect spies, every spy on node/.append style={thick}}] 
			\node {\pgfimage[width=0.45\linewidth]{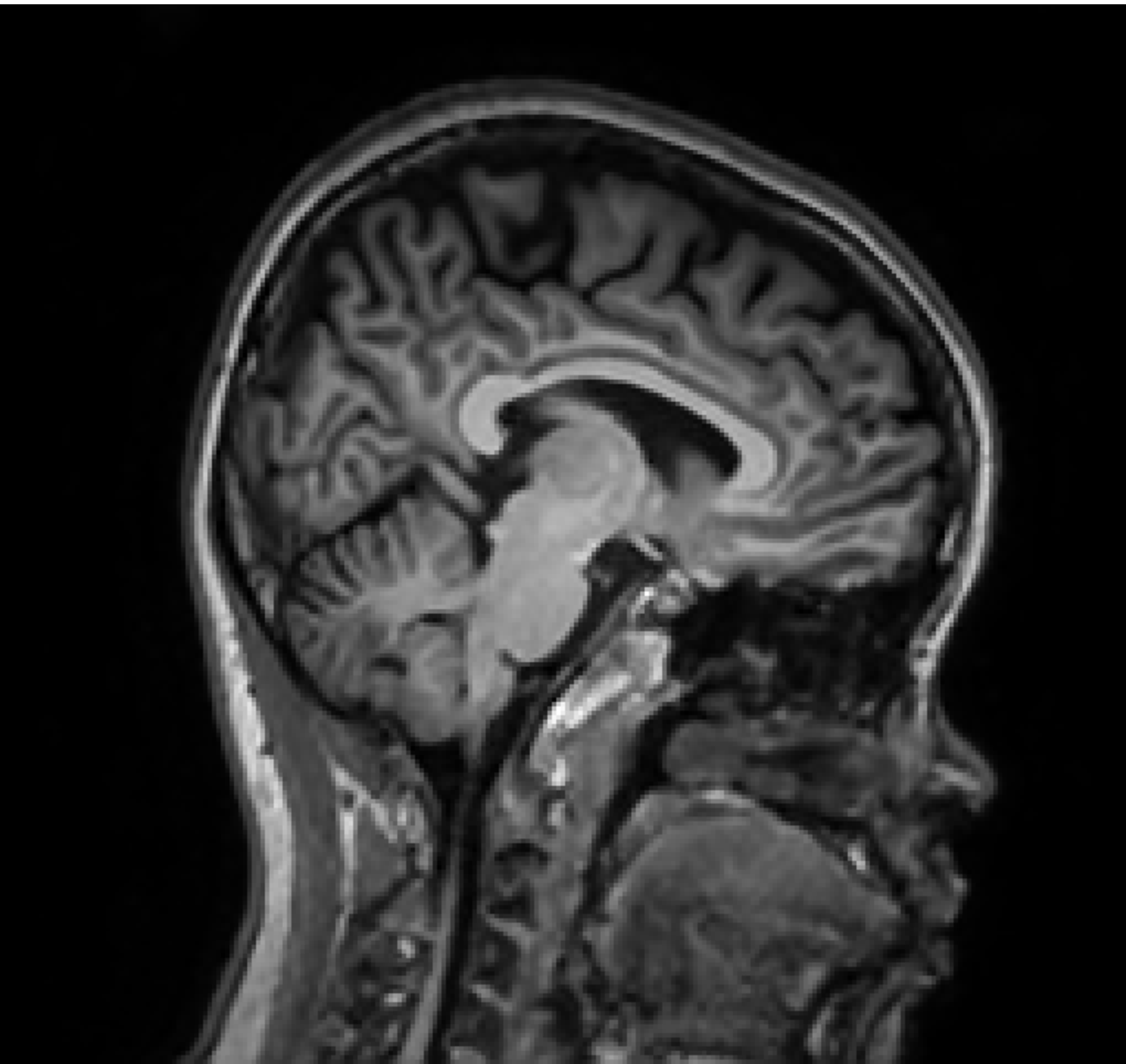}}; 
      		\spy on (-0.70, 0) in node [left] at (1.915, -0.895); 
    	\end{tikzpicture}
	}\hspace{-1.3em}
	}
	\vspace{-0.8em}
	\caption{Our \textit{volumetric super-resolution forests} (VSRF) facilitate high-resolution MRI while retaining a low scan time as illustrated by one sagittal slice of the Kirby 21 MRI dataset \cite{Landman2011} (SR factor 2).}
	\label{fig:kirby}
\end{figure}
\setlength{\textfloatsep}{5pt plus 1.0pt minus 2.0pt}

SISR techniques in 3-D MRI can be either used on \textit{slice} or \textit{volume} level. Slice-wise methods enhance the resolution within one plane but ignore the coherence between adjacent slices of volumetric data.
Volumetric methods enable simultaneous resolution enhancement in all directions.
For instance, Manj{\'o}n et al. \cite{Manjon2010} applied an iterative patch-based non-local reconstruction scheme based on self-similarity and the 3-D non-local means filter.
Jog et al. \cite{Jog2016} build up on anchored neighborhood regression \cite{Timofte2013} to perform self SR.
A closely related field to MRI SR, image synthesis, uses dictionaries \cite{Roy2013} or random forests \cite{Jog2014} to super-resolve MR volumes by utilizing additional information of a HR volume with another MR contrast of the same subject. Alexander et al. \cite{Alexander2014} implemented a random forest for SR in diffusion MRI with cubic patches 
and features customized for diffusion tensor images (DTIs).
Yoldemir et al. \cite{Yoldemir2014} applied dictionary learning to diffusion-weighted 3-D images for volumetric SR. Recently, Tanno et al. \cite{Tanno2017} integrated an uncertainty modeling with a 3-D convolutional neural network (CNN) on DTIs. 
Bahrami et al. \cite{Bahrami2016} used a CNN with anatomical and appearance features to non-linearly map 3T to 7T MR volumes. In a further work \cite{Bahrami2017}, they estimated a 7T-like volume with a random forest based on 3T and 7T patches and enhanced the 7T-like volume with dictionary learning. 
For brain MRI, Rueda et al. \cite{Rueda2013} applied dictionary learning for 3-D SR. More recently, Pham et al. \cite{Pham2017} and Chen et al. \cite{Chen2018} proposed deep learning by extending 2-D CNNs for natural images to 3-D CNNs. 
CNNs avoid the need for hand-crafted features and enable end-to-end learning. This leads to state-of-the art results but requires large amounts of training data that need to match to the desired application to be effective \cite{Mousavi2018}. 
In contrast, random forests yield robust results with only a small amount of training data and are very fast to train even without sophisticated hardware.

In this paper, we propose \textit{volumetric SR forests} (VSRF) for example-based resolution enhancement in MRI. Our method employs random forest regression and generalizes SR forests (SRF) \cite{Schulter2015} that have been originally introduced for natural images to volumetric data. We train the random forest with overlapping 3-D patches and new \textit{customized features}. For robust regression, we propose a \textit{median ensemble model} to obtain the forest prediction. 
Our proposed VSRF can effectively improve the spatial resolution in MRI, see Fig.~\ref{fig:kirby}. It facilitates computationally efficient training and inference even on a CPU. Moreover, it is effective for example-based SR even with a limited amount of training data making it an attractive tool for practical applications.

\section{Volumetric Super-Resolution Forests}
\label{sec:vsrf}
Our proposed VSRF builds on random forest regression \cite{Schulter2015} to learn a locally linear mapping between LR and HR patches from example data. 
We introduce training, inference and features for this model.
\subsection{Random Forest Training}
\label{ssec:trainRF}
A random forest consists of a set of decision trees which are independently trained on a training set of $N$ patch pairs $\{ \mathbf{x}_L^n, \mathbf{x}_H^n \}_{n=1}^N$, where $\mathbf{x}_L \in \mathbb{R}^{D_L}$ and $\mathbf{x}_H \in \mathbb{R}^{D_H}$ are feature vectors representing the LR and HR patches. The feature vectors $\mathbf{x}_L^n$ and $\mathbf{x}_H^n$ are stacked into matrices $\mathbf{X}_L \in \mathbb{R}^{D_L \times N}$ and $\mathbf{X}_H \in \mathbb{R}^{D_H \times N}$ \cite{Schulter2015}. 
The tree structure is learned by recursively dividing the data space into disjoint subsets until the maximum tree depth or minimum number of feature vector pairs in a node is reached and a leaf node is created. The splits are performed based on a binary decision made at each internal node using a pair-wise difference splitting function \cite{Schulter2015, Dollar2013}:
\begin{equation}
h(\mathbf{x}_L,\boldsymbol{\theta}) = 
\begin{cases}
1 & \mathbf{x}_L[\varphi_1] - \mathbf{x}_L[\varphi_2] < \tau,\\
0 &\text{otherwise}
\end{cases}
\end{equation} 
which compares the difference of two randomly selected feature dimensions $\varphi_1, \varphi_2 \in \{ 1, \ldots, D_L\}$ from the LR feature vector $\mathbf{x}_L$ to a threshold $\tau$. 
The parameters $\boldsymbol{\theta} = \{\varphi_1, \varphi_2, \tau \} $ for the splitting functions are estimated by evaluating a quality measure. 
We use node optimization for finding the splitting parameters, i.e. we do not evaluate the quality measure on all training samples of the node but we randomly subsample the data of that node \cite{Schulter2015}.
Thus, we reduce computation time and increase the variation between the trees. 
A common choice for the quality measure for regression forests is reduction-in-variance based on information gain \cite{Bahrami2017,Dollar2013}. We use a modified quality measure defined by Schulter et al. \cite{Schulter2015} that operates in both the low and high-resolution space according to:
\begin{equation}
Q(\mathbf{X}_H, \mathbf{X}_L, \boldsymbol{\theta}) = \sum_{i \in \{left,right\}} N_i \cdot E(\mathbf{X}_H^i,\mathbf{X}_L^i),
\end{equation}
where $\mathbf{X}_L^i$ and $\mathbf{X}_H^i$ contain the $N_i$ LR and HR feature vectors of the left and right subsets according to $\boldsymbol{\theta}$. 
The optimization of the quality measure selects the splits for which the variance 
of HR feature vectors and variance of LR feature vectors in the subsets is minimal. The variance for both domains \cite{Schulter2015} is given by:
\begin{equation}
E(\mathbf{X}_H^i,\mathbf{X}_L^i) = \frac{1}{N_i} \sum_{n=1}^{N_i} ( \| \mathbf{x}_H^n - \bar{\mathbf{x}}_H^n \|_2^2 + \kappa \cdot \| \mathbf{x}_L^n - \bar{\mathbf{x}}_L^n \|_2^2 ),
\end{equation}
where $\bar{\mathbf{x}}$ is the mean over all samples and $\kappa$ is a hyper-parameter to control the influence of the LR variance.

In the leaf nodes, we learn locally linear mappings from LR feature vectors to HR feature vectors using the feature vector pairs that reach these specific nodes. We determine a mapping $\hat{\mathbf{W}}_l$ for the leaf $l$ according to the least squares problem \cite{Schulter2015}:
\begin{equation}
	\hat{\mathbf{W}}_l = \argmin_{\mathbf{W}_l} \sum_{n=1}^{N_l} \| \mathbf{x}_H^n - \mathbf{W}_l \cdot \mathbf{x}_L^n  \|_2^2.
\end{equation}
To yield a more stable solution, we use ridge regression:
\begin{equation}
 \hat{\mathbf{W}}_l = \argmin_{\mathbf{W}_l} \| \mathbf{X}_H^l - \mathbf{W}_l \mathbf{X}_L^l  \|_2^2 + \lambda \| \mathbf{W}_l \|_2^2,
\end{equation}
where $\lambda$ is the regularization parameter. The estimate for the linear mapping can be solved in closed-form \cite{Schulter2015}:
\begin{equation}
\hat{\mathbf{W}}_l^\top = ( \mathbf{X}_L^{l \top} \mathbf{X}_L^l + \lambda \mathbf{I} )^{-1} \mathbf{X}_L^{l \top} \cdot \mathbf{X}_H^l. 
\end{equation}
\subsection{Random Forest Inference}
\label{ssec:inferenceRF}
In random forest inference the LR feature vectors of the test set are sent through all trees resulting in one prediction of the mapping function for each LR feature vector and each tree. The estimated predictions are then combined to a single HR patch by the forest ensemble model. A common approach for ensembles is to average the prediction of all trees. For SR this means computing an element-wise average of the predicted HR patches. To account for outliers of the predicted values of the trees, we employ a median ensemble model which is more robust with regards to these issues and computes the median for each component of the predicted HR patches.
The predicted HR information is then added to the tricubically upsampled LR volume. Since we extracted overlapping patches, we reconstruct the final volume by averaging the overlapping voxels.
\subsection{Features and Patch Extraction}
\label{ssec:featandpatch}
In the proposed VSRF method, LR volumes are processed in two steps:
1) The LR volume $\mathbf{V}_L$ is upscaled to the size of the target HR volume by tricubic interpolation resulting in the volume $\tilde{\mathbf{V}}_L$.
\mbox{2) The} missing high-frequency information is predicted by the random forest and added to $\tilde{\mathbf{V}}_L$.
In order to perform 3-D SR, we extract overlapping $n_x \times n_y \times n_z$ patches from $\tilde{\mathbf{V}}_L$. 
We extract corresponding HR patches of equal size from the difference volume $\mathbf{V}_H - \tilde{\mathbf{V}}_L$ that contains the missing high-frequency information of $\tilde{\mathbf{V}}_L$ compared to the HR volume $\mathbf{V}_H$.

Based on the LR patches, we compute a set of ten different features to which we refer as DevEdge in the following: 
Partial first- and second-order derivatives in all three directions (Dev), edge orientation in all three directions and edge magnitude. First- and second-order derivatives are used in the SR framework of \cite{Timofte2013} for \mbox{2-D} images, which we extend to 3-D by adding a third component in $z$-direction. 
Edge magnitude $M$ and edge orientation $\phi$ are computed based on the first-order derivatives $D_i$ with $i \in \{x,y,z\}$:
\begin{equation}
M = \sqrt{\smash[b]{D_x^2 + D_y^2 + D_z^2}}, \quad \phi_{xy} = \arctan \left( \frac{D_y}{D_x} \right), 
\end{equation}
the edge orientations $\phi_{zx}$ and $\phi_{zy}$ are computed accordingly. 
Since edge orientation is very sensitive to noise, we convolve 
$\tilde{\mathbf{V}}_L$ with a Gaussian filter with standard deviation $\sigma$ before we compute the derivatives for edge orientation and magnitude.
\begin{table*}[t]  
  \footnotesize
	\centering
	\caption{Mean PSNR (dB) and SSIM on the mouse brain dataset (5 volumes) and on the Kirby 21 human brain dataset \cite{Landman2011} (30 volumes). We compared the proposed volumetric SR forest (VSRF) to different state-of-the-art SR methods (SR factor 2).}
	\setlength\tabcolsep{2.3pt} 
	\begin{tabular}{*{24}{c}}
		\toprule
		\textbf{Dataset} & \multicolumn{2}{c}{Tricubic} & & \multicolumn{2}{c}{2-D SRF \cite{Schulter2015}} & & \multicolumn{2}{c}{Psd. 3-D SRCNN} &  & \multicolumn{2}{c}{Psd. 3-D SRF} & & \multicolumn{2}{c}{NLMU \cite{Manjon2010}} & & \multicolumn{2}{c}{VANR} & & \multicolumn{2}{c}{VA+} &  &\multicolumn{2}{c}{VSRF (ours)}\\
\cmidrule{2-24} 	 
		 & PSNR & SSIM  & & PSNR & SSIM & & PSNR & SSIM & & PSNR & SSIM & & PSNR & SSIM & & PSNR & SSIM & & PSNR & SSIM & & PSNR & SSIM \\
		 \midrule
		\textbf{Mouse Brain} & 34.94 & 0.9637 & & 37.31 & 0.9752 & & 36.82 & 0.9680 & & 38.63 & 0.9781 & & 36.94 & 0.9721 & & 37.69 & 0.9750 & & 38.75 & 0.9779 & & \textbf{39.46} & \textbf{0.9804} \\ 		
		\midrule
		\textbf{Kirby Brain} & 34.84 & 0.9502 & & 35.58 & 0.9611 & & 36.10 & 0.9643 & & 36.48 & 0.9659 & & 36.58 & 0.9662 & & 35.59 & 0.9605 & & 36.06 & 0.9650 & & \textbf{37.15} & \textbf{0.9701} \\ 
		\bottomrule
	\end{tabular}
	\label{tab:res01}
\end{table*}
To accelerate the training process, a PCA dimensionality reduction is applied to the feature vectors like in \cite{Timofte2013}.
These feature vectors together with the vectorized HR patches are then used as input for the random forest. 
\section{Experiments and Results}
\label{sec:experiments}
We evaluated the performance of VSRF with a mouse and a human brain MRI dataset.
The mouse brain MRI dataset consisted of 21 isotropic high-resolution volumes\footnote{All animal experiments were approved by the local ethic committee (Regierung von Unterfranken, W\"urzburg, Germany)}. 
The MR volumes were acquired with a 4.7T MR scanner (Bruker BioSpec 47/40) with $0.05 \times 0.05 \times 0.05\, \mathrm{mm}^3$ resolution, which were measured in-situ after transcardial perfusion with ProHance\textsuperscript{TM} with a T1-weighted \mbox{3-D} FLASH sequence ($\mathit{TE} = 4.6\, \mathrm{ms}$, $\mathit{TR} = 16.3\, \mathrm{ms}$, field of view (FOV): $17\, \mathrm{mm} \times 15\, \mathrm{mm} \times 8\, \mathrm{mm}$, matrix: $340 \times 300 \times 160$ voxels, flip angle $= 20 \,^{\circ}$, $\mathit{NEX} = 14$, total scan time $ = 3\, \mathrm{h} \, 2 \, \mathrm{min} \, 37 \mathrm{s}$).
We manually segmented the mouse brain and cropped the volumes to $320 \times 210 \times 140$ voxels. 
The LR volumes were generated by tricubic downscaling of the HR volumes by an isotropic scaling factor of 2. 
For LR volumes the acquisition time would be reduced substantially by a factor of $5.58$ to $32\, \mathrm{min}\, 42 \,\mathrm{s}$ at $\mathit{NEX} =10$ keeping comparable signal-to-noise ratio (SNR). 
We used 13 pairs of LR and HR volumes for training, three to validate our choice of parameters and five for testing.
Second, we applied the SR methods to the publicly available Kirby 21 dataset \cite{Landman2011} with human brain MR data. 
We used the 42 T1-weighted MPRAGE volumes with $1.0 \times 1.0 \times 1.2\, \mathrm{mm}^3$ resolution, which were acquired with a 3T MR scanner (Achieva, Philips Healthcare) in the sagittal plane
and we cropped them to $240 \times 204 \times 256$ voxels. The LR volumes were generated by tricubic downscaling by a factor of 2. The 42 pairs of HR and LR volumes were divided into a training set of ten (No. 33-42), a validation set of two (No. 31-32) and a test set of 30 volume pairs (No. 1-30).

\textbf{Comparison to State-of-the-Arts.}
For both datasets, we compared the performance of VSRF with tricubic upsampling, 3-D SR methods, pseudo 3-D SR methods and the 2-D SRF \cite{Schulter2015}.
SRF was applied slice-wise to the volumetric MR data using $3 \times 3$ patches, $T = 30$ trees, $\kappa = 1$ and $\lambda$ was estimated automatically (according to the source code of \cite{Schulter2015}) based on the condition number of the least squares problem.
The parameter settings for our VSRF are: $T = 30$ trees, $3 \times 3 \times 3$ patches, $\kappa = 1$ and automatic $\lambda$ and $\sigma = 1$ for the edge orientation features. 
For 3-D SR, we extended the 2-D anchored neighborhood regression (ANR) \cite{Timofte2013} and adjusted anchored neighborhood regression (A+) \cite{Timofte2015} by employing 3-D patches and utilizing the same features as for VSRF. We refer to these volumetric methods as VANR and VA+. For VANR and VA+ we used $3 \times 3 \times 3$ patches, a dictionary size of $2048$ and neighborhoods of $16$ atoms for VANR and $2048$ samples for VA+. We directly utilized all training samples for the computation of VA+ regressors instead of augmenting the samples by means of a scaling pyramid and then restricting the number to 5 million like in \cite{Timofte2015}.
Also we applied the non-local MRI upsampling (NLMU) \cite{Manjon2010} as a 3-D SR method.
In addition, we extended the \mbox{2-D} SRF \cite{Schulter2015} and 2-D SRCNN \cite{Dong2014} to pseudo 3-D SRF and pseudo 3-D SRCNN. 
Following prior work \cite{Pham2017}, the pseudo \mbox{3-D} methods were
constructed by averaging the three slice-wise SRFs or SRCNNs in axial, sagittal and coronal view of the MR volume. 
We trained the 2-D SRCNN from scratch with an augmented MRI dataset over 15 million iterations (1383 epochs for the mouse and 1633 epochs for the Kirby 21 dataset) by feeding the SRCNN with slices of the axial, sagittal and coronal view of the volume.

The SR results on the mouse and Kirby 21 dataset were evaluated using the peak signal-to-noise ratio (PSNR) and the structural similarity (SSIM) \cite{Wang2004}, see Table~\ref{tab:res01}.
VSRF achieved the highest PSNR and SSIM of all tested methods on both datasets. 
On the mouse dataset VSRF achieved a gain in PSNR (SSIM) of $0.71\,\mathrm{dB}\,(0.0025)$ compared to VA+ and of $0.83\,\mathrm{dB}\,(0.0023)$ compared to pseudo \mbox{3-D} SRF (next best results).
On the Kirby 21 dataset NLMU performed second best with a $0.57 \,\mathrm{dB}\,(0.0039)$ lower PSNR (SSIM) than VSRF.
Compared to tricubic upsampling VSRF increased the PSNR (SSIM) by $4.52\,\mathrm{dB}\,(0.0167)$ on the mouse and by $2.32\,\mathrm{dB}\,(0.0199)$ on the Kirby dataset. 
%
%
\begin{figure*}[tb]
\begin{minipage}[b]{.19\linewidth}
  \centering
  \centerline{\includegraphics[width=3.2cm]{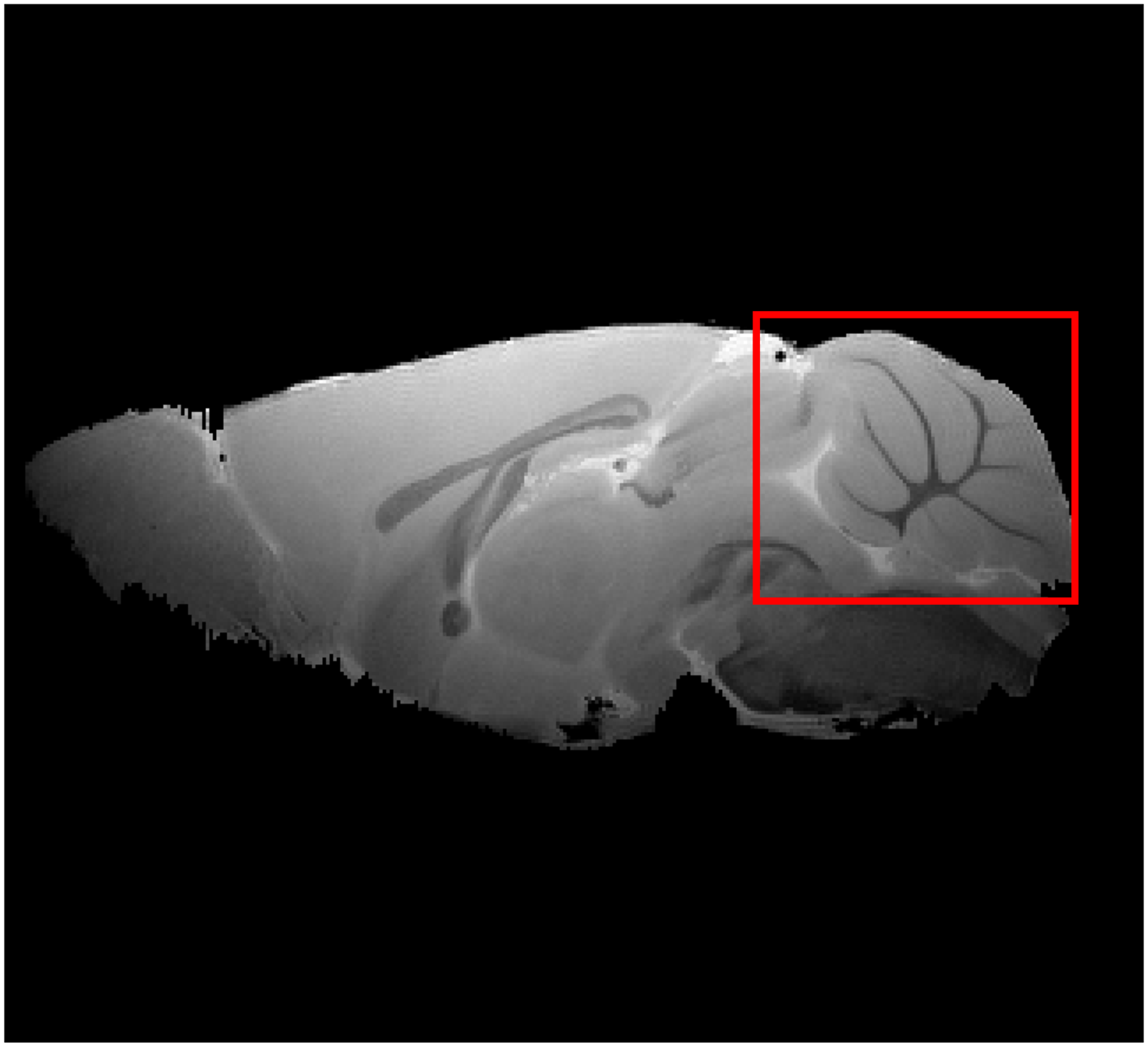}}
    \centerline{(a) ROI Ground Truth}\medskip
\end{minipage}
\hfill
\begin{minipage}[b]{.19\linewidth}
  \centering
  \centerline{\includegraphics[width=3.2cm]{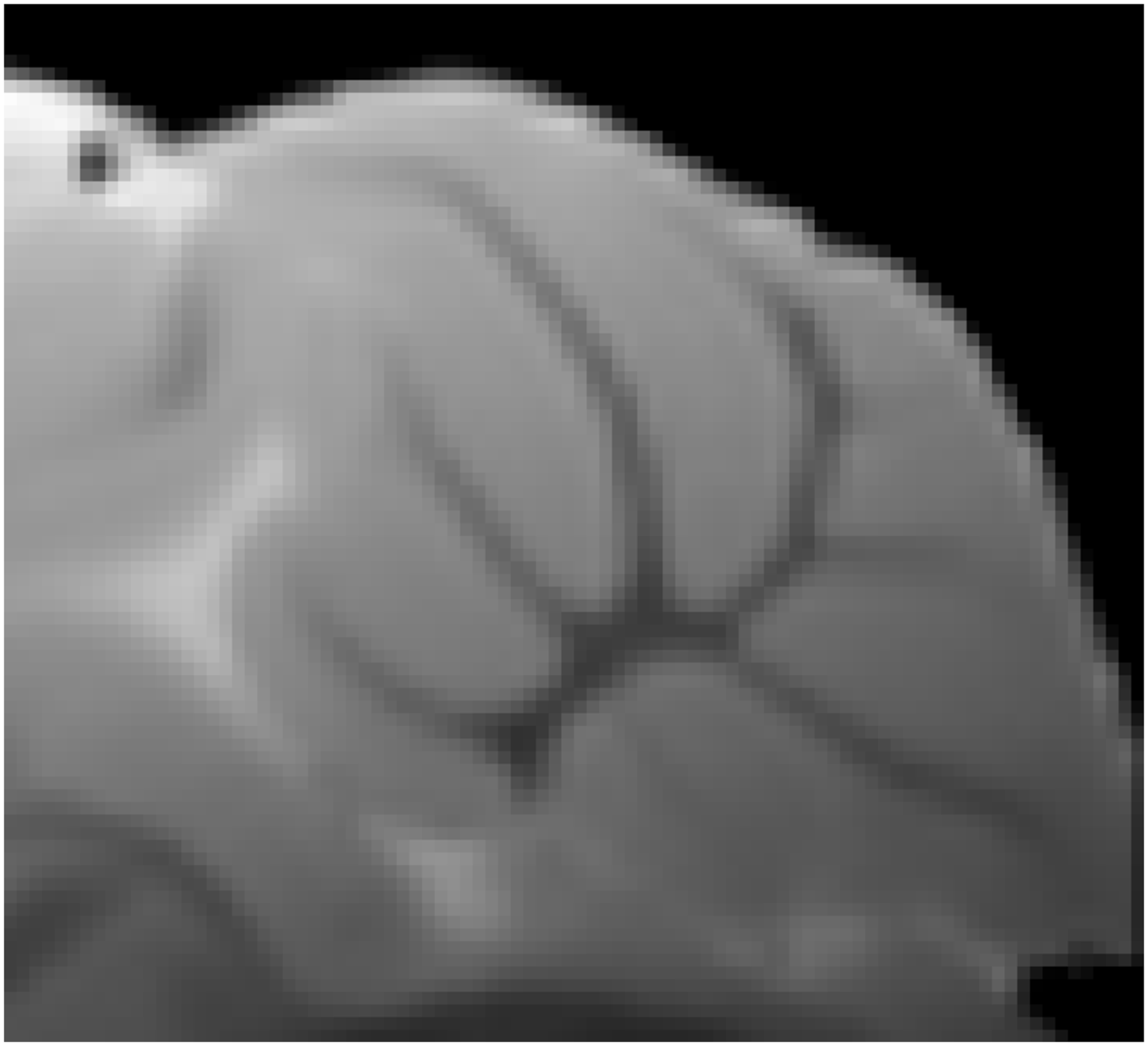}}
  \centerline{(b) Tricubic}\medskip
\end{minipage}
\hfill
\begin{minipage}[b]{.19\linewidth}
  \centering
  \centerline{\includegraphics[width=3.2cm]{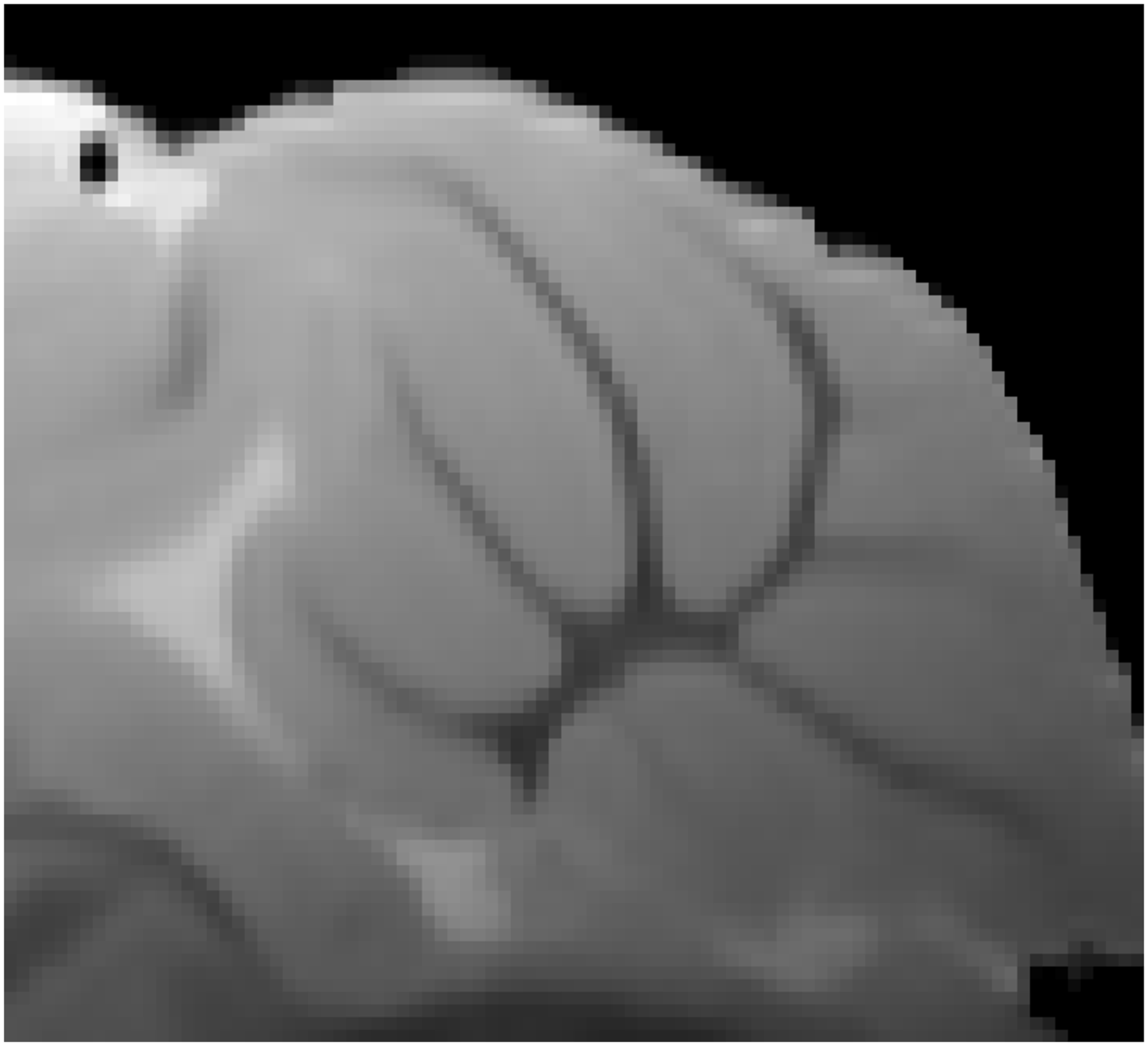}}
  \centerline{(c) SRF \cite{Schulter2015}}\medskip
\end{minipage}
\hfill
\begin{minipage}[b]{.19\linewidth}
  \centering
  \centerline{\includegraphics[width=3.2cm]{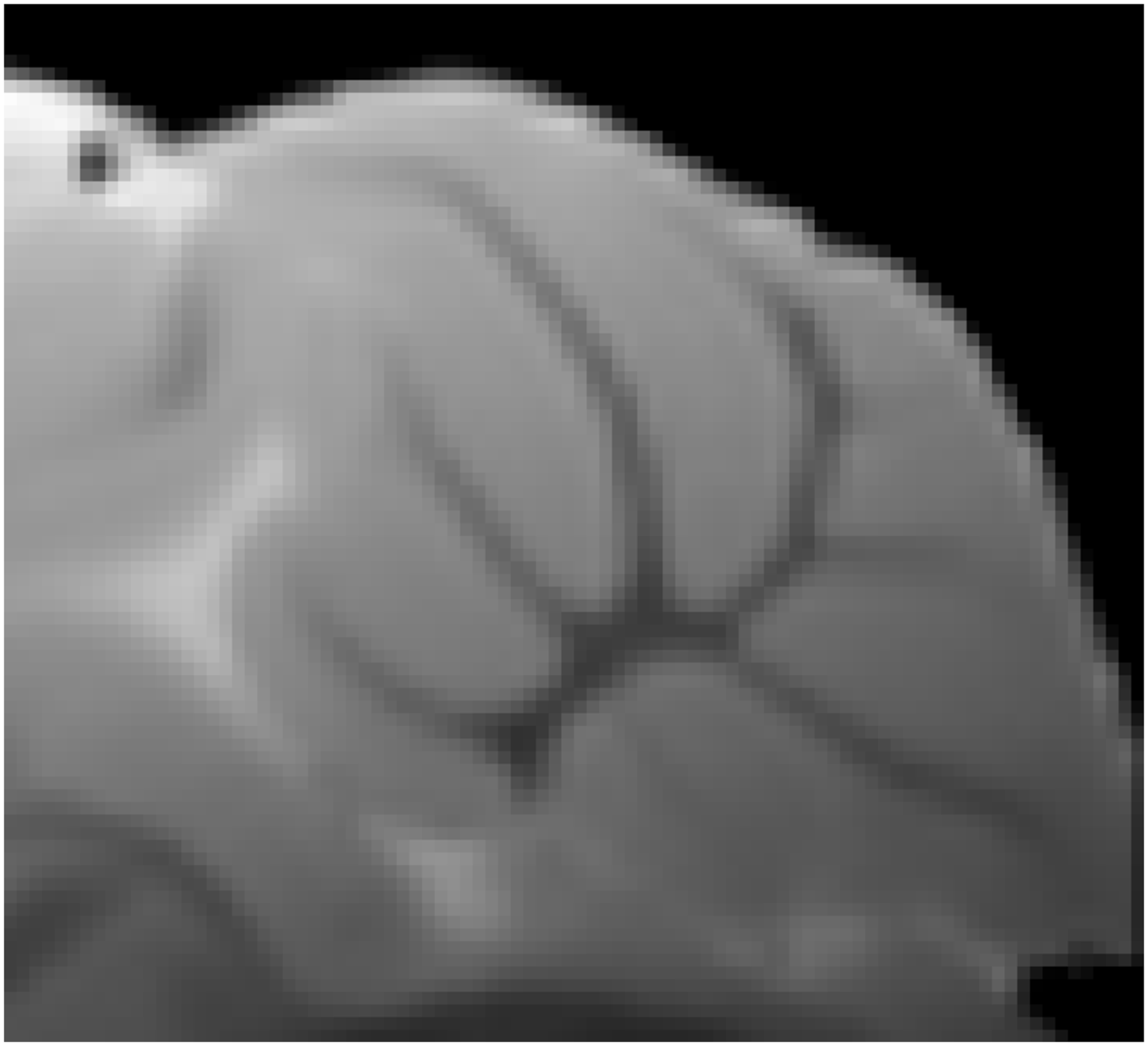}}
  \centerline{(d) Pseudo 3-D SRCNN}\medskip
\end{minipage}
\hfill
\begin{minipage}[b]{.19\linewidth}
  \centering
  \centerline{\includegraphics[width=3.2cm]{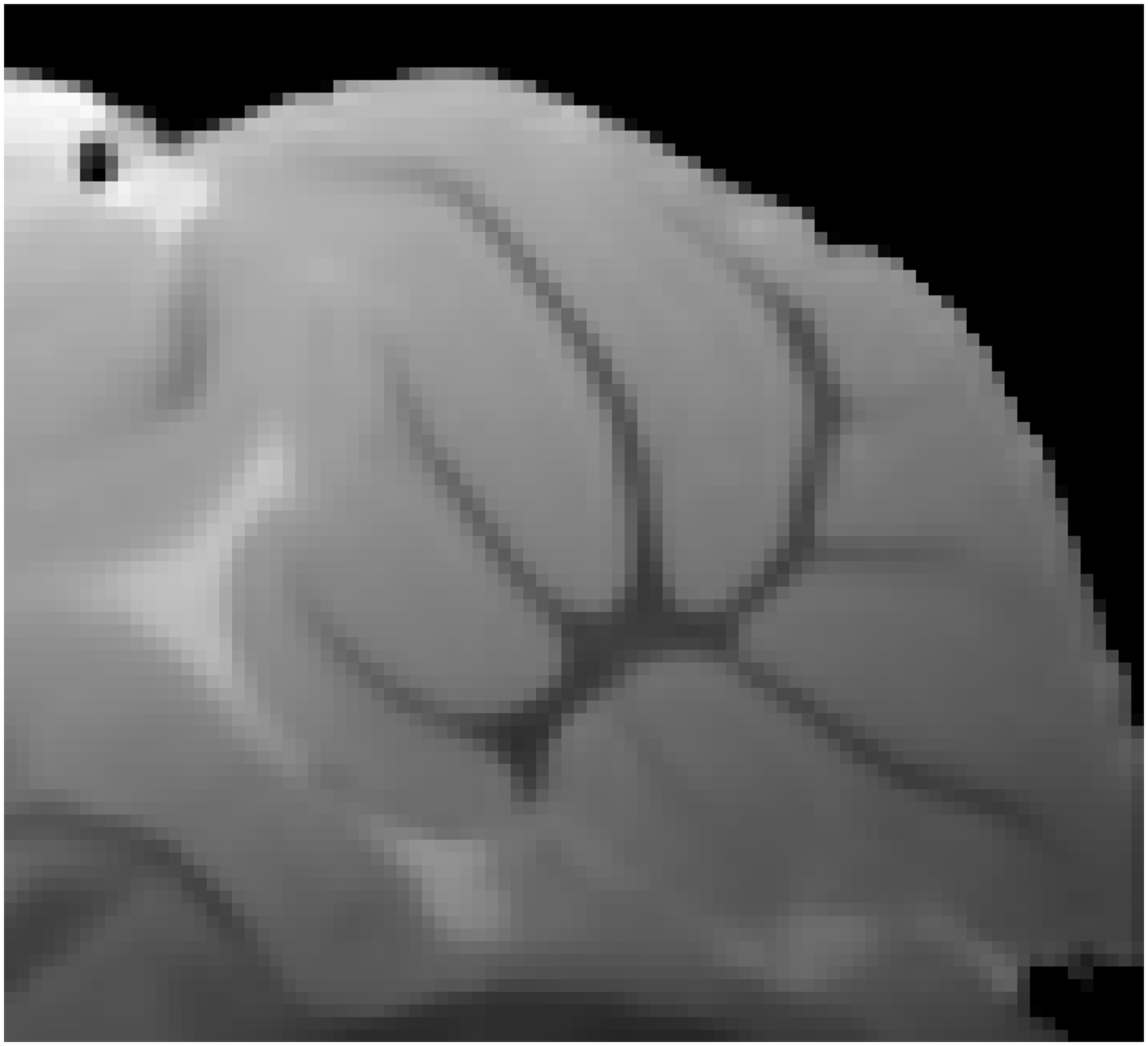}}
  \centerline{(e) Pseudo 3-D SRF}\medskip
\end{minipage}
\begin{minipage}[b]{.19\linewidth}
  \centering
  \centerline{\includegraphics[width=3.2cm]{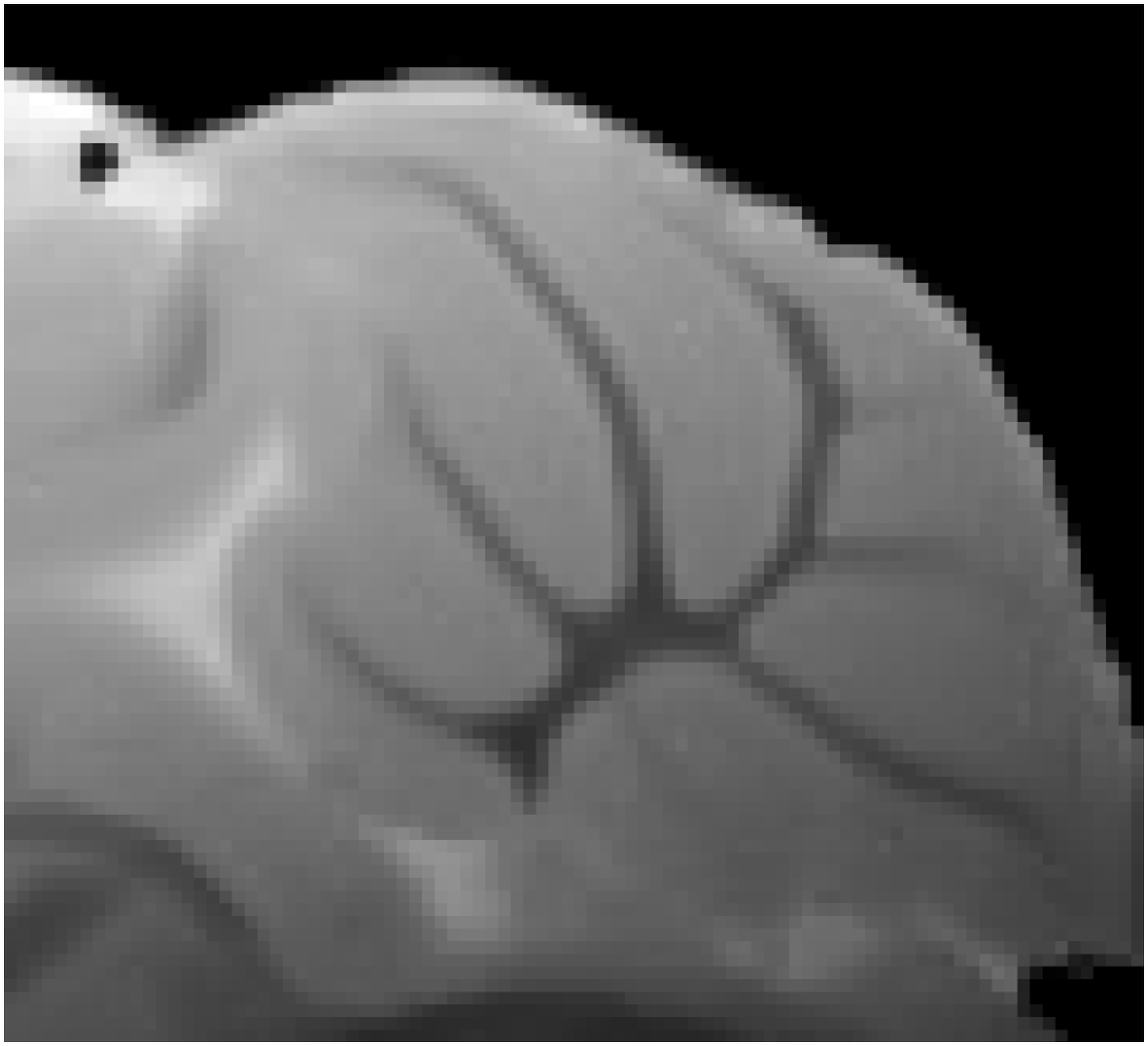}}
  \centerline{(f) NLMU \cite{Manjon2010}}\medskip
\end{minipage}
\hfill
\begin{minipage}[b]{.19\linewidth}
  \centering
  \centerline{\includegraphics[width=3.2cm]{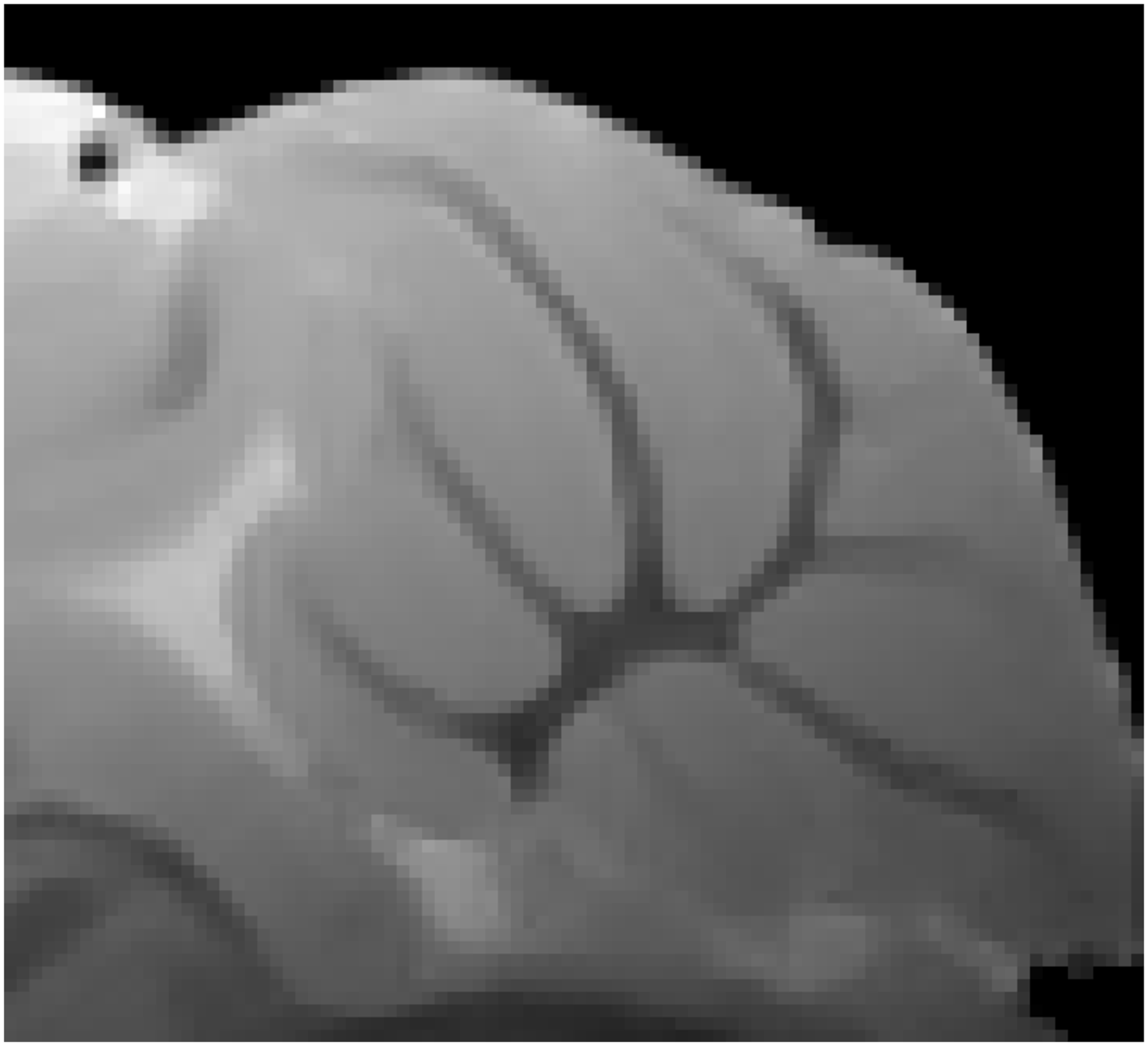}}
  \centerline{(g) VANR}\medskip
\end{minipage}
\hfill
\begin{minipage}[b]{.19\linewidth}
  \centering
  \centerline{\includegraphics[width=3.2cm]{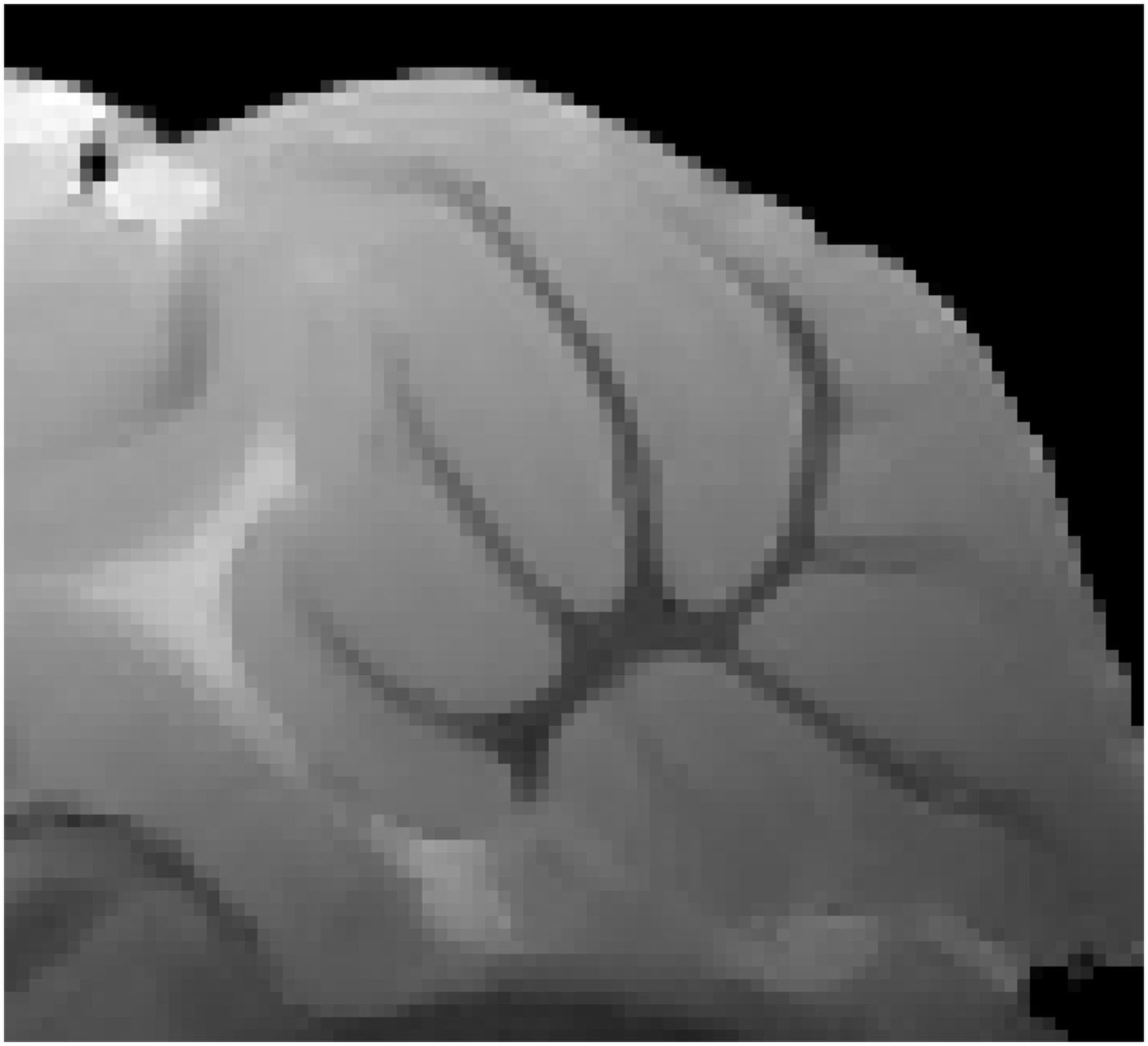}}
  \centerline{(h) VA+}\medskip
\end{minipage}
\hfill
\begin{minipage}[b]{.19\linewidth}
  \centering
  \centerline{\includegraphics[width=3.2cm]{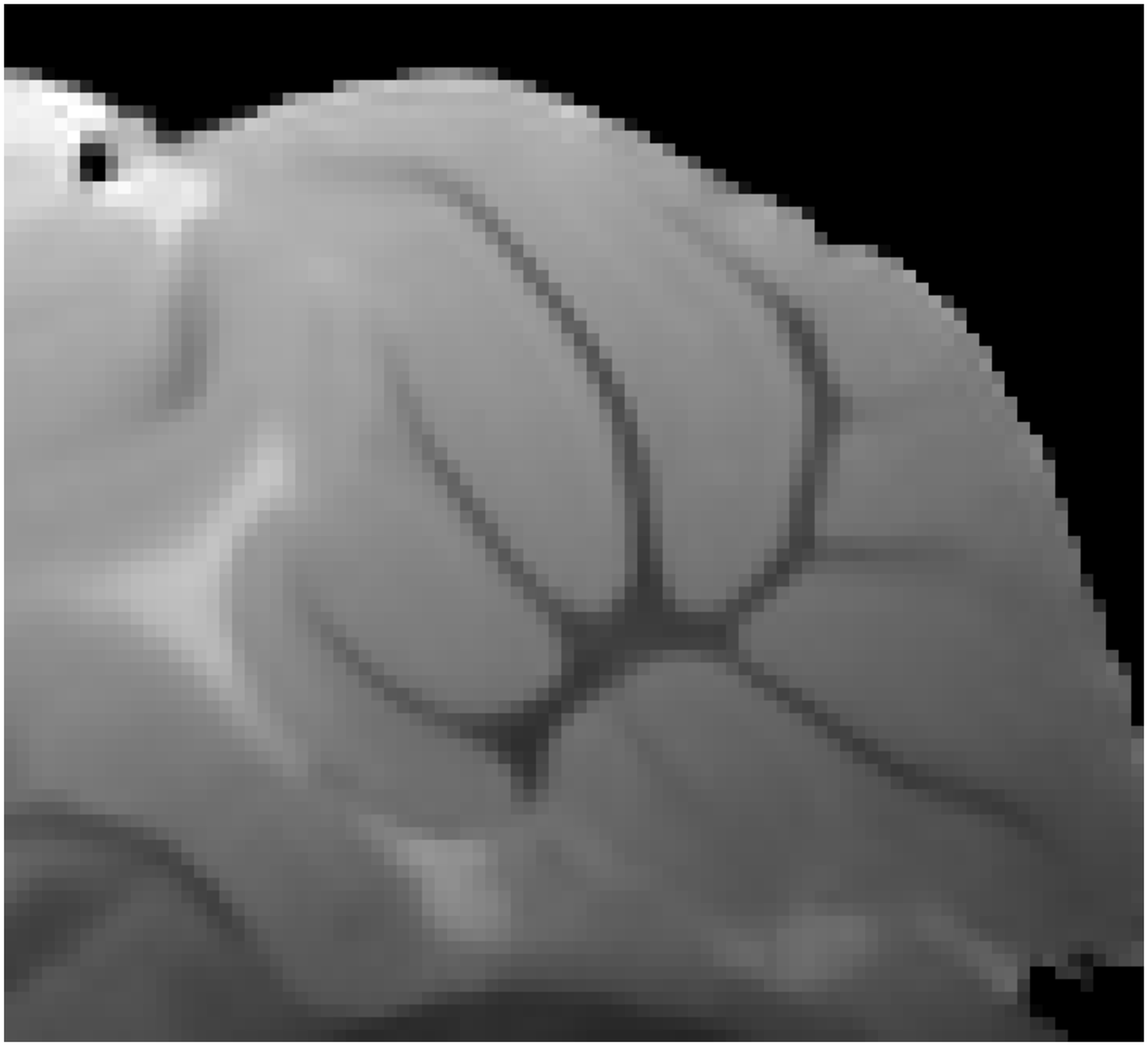}}
  \centerline{(i) VSRF (ours)}\medskip
\end{minipage}
\hfill
\begin{minipage}[b]{.19\linewidth}
  \centering
  \centerline{\includegraphics[width=3.2cm]{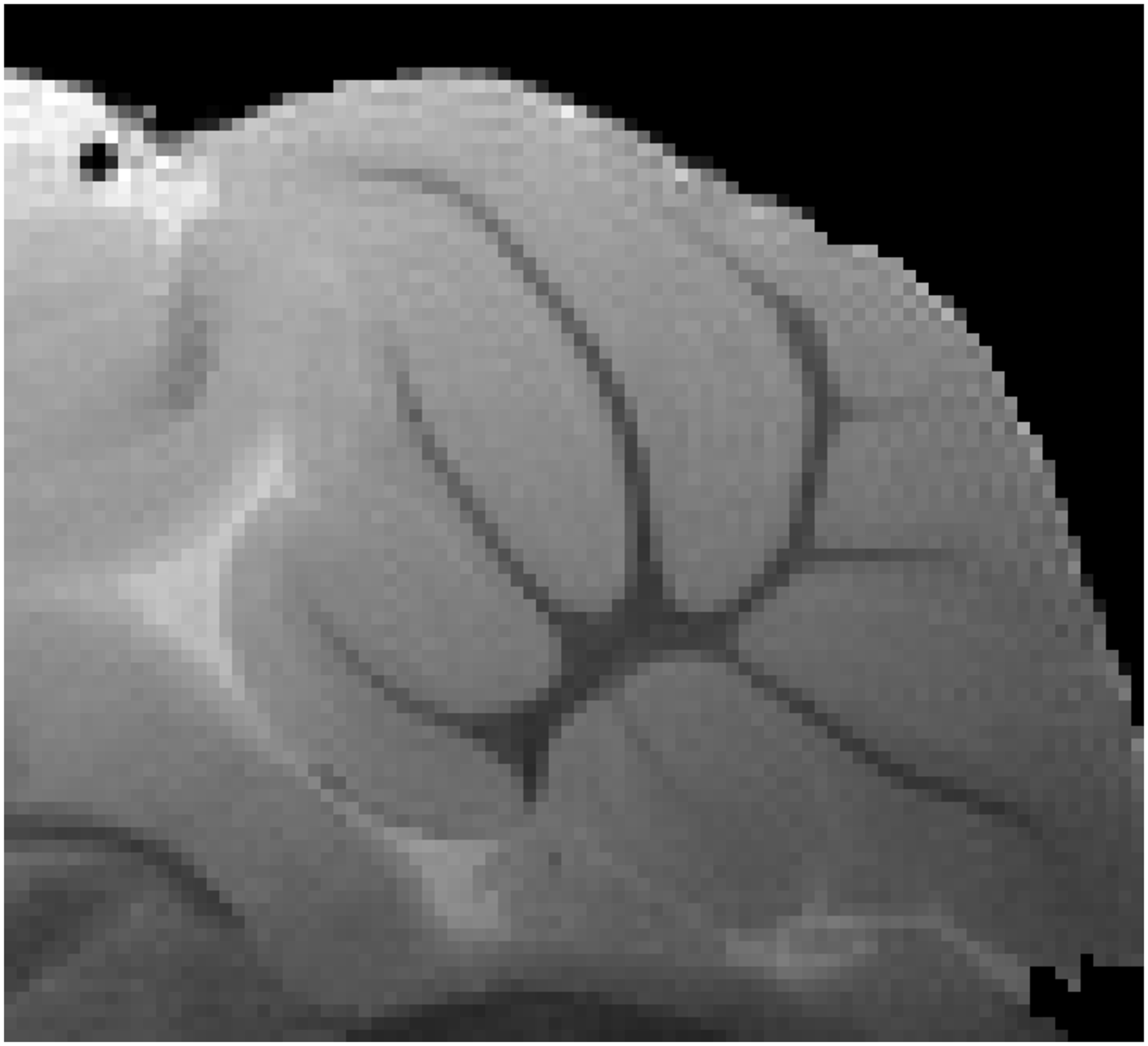}}
  \centerline{(j) Ground Truth (GT)}\medskip
\end{minipage}
\vspace{-0.8em}
\caption{Visual comparison of SR results for one sagittal slice of the mouse brain MRI dataset (SR factor 2).}
\label{fig:mouseSagittal}
\end{figure*}
%
\begin{figure*}[htb]
\begin{minipage}[b]{.135\linewidth}
  \centering
  \centerline{\includegraphics[width=2.4cm]{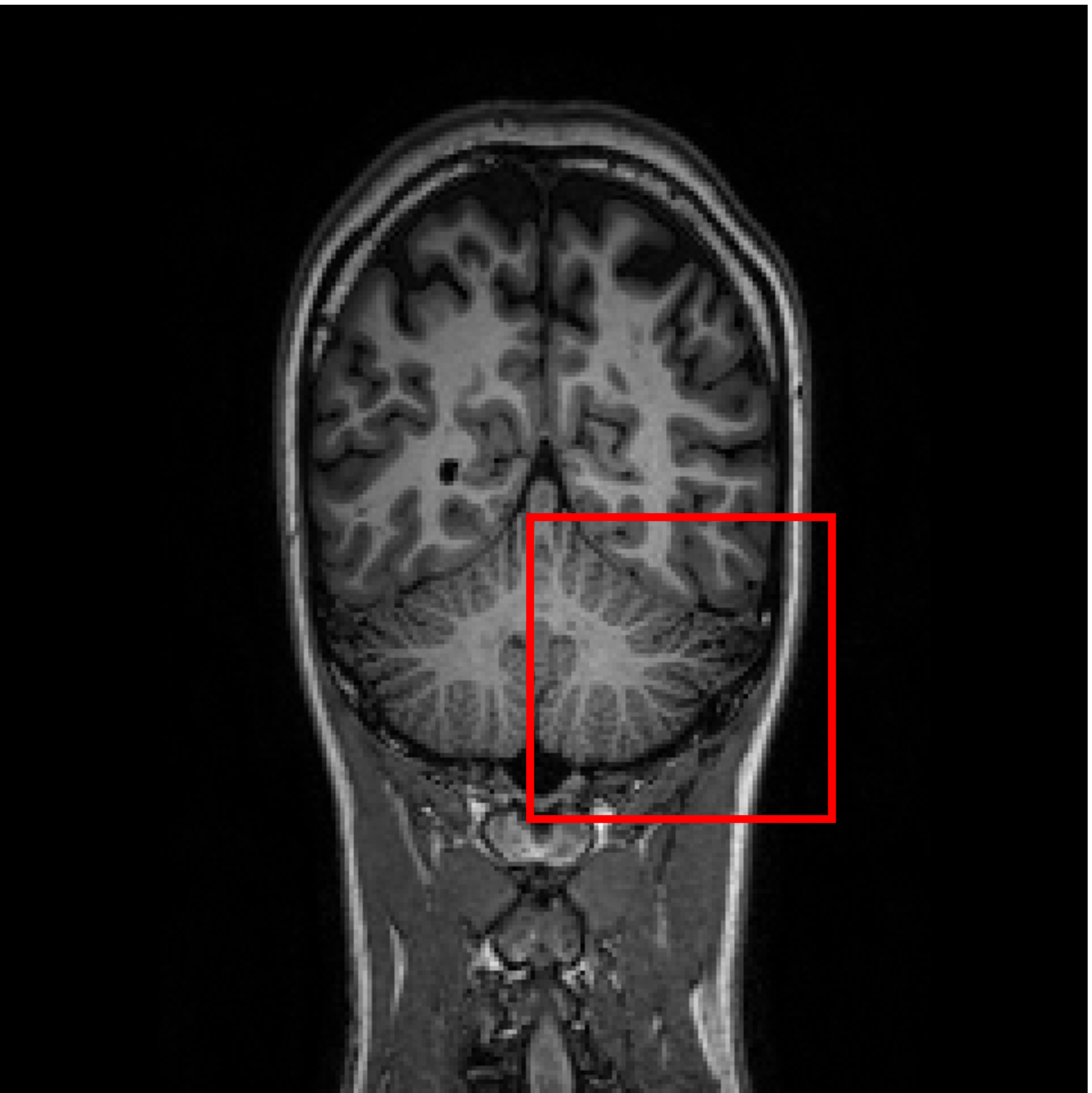}}
    \centerline{(a) ROI GT}\medskip
\end{minipage}
\hfill
\begin{minipage}[b]{.135\linewidth}
  \centering
  \centerline{\includegraphics[width=2.4cm]{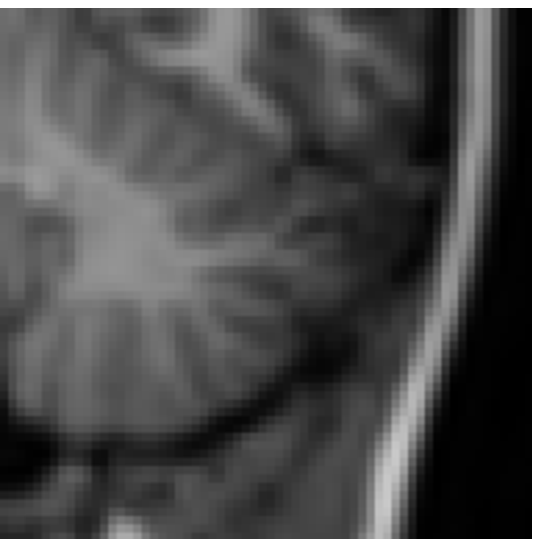}}
  \centerline{(b) Tricubic}\medskip
\end{minipage}
\hfill
\begin{minipage}[b]{.135\linewidth}
  \centering
  \centerline{\includegraphics[width=2.4cm]{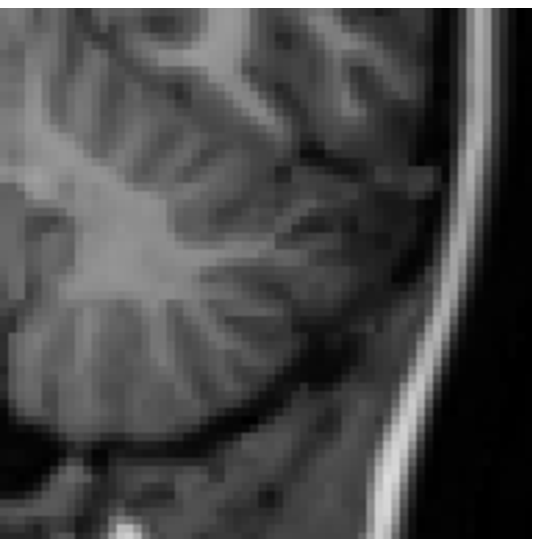}}
  \centerline{(c) NLMU \cite{Manjon2010}}\medskip
\end{minipage}
\hfill
\begin{minipage}[b]{.135\linewidth}
  \centering
  \centerline{\includegraphics[width=2.4cm]{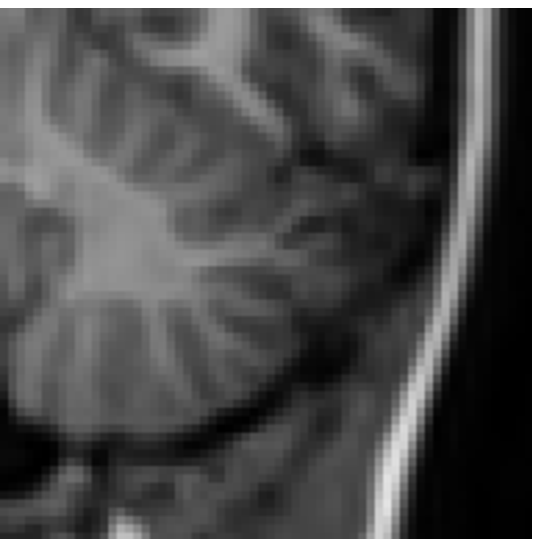}}
  \centerline{(d) VANR}\medskip
\end{minipage}
\hfill
\begin{minipage}[b]{.135\linewidth}
  \centering
  \centerline{\includegraphics[width=2.4cm]{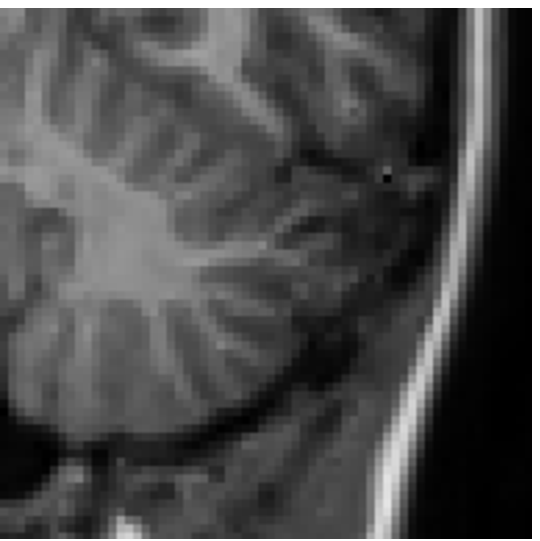}}
  \centerline{(e) VA+}\medskip
\end{minipage}
\hfill
\begin{minipage}[b]{.135\linewidth}
  \centering
  \centerline{\includegraphics[width=2.4cm]{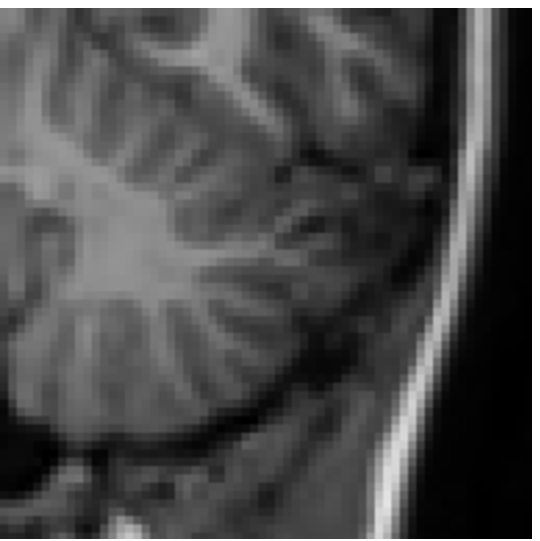}}
  \centerline{(f) VSRF (ours)}\medskip
\end{minipage}
\hfill
\begin{minipage}[b]{.135\linewidth}
  \centering
  \centerline{\includegraphics[width=2.4cm]{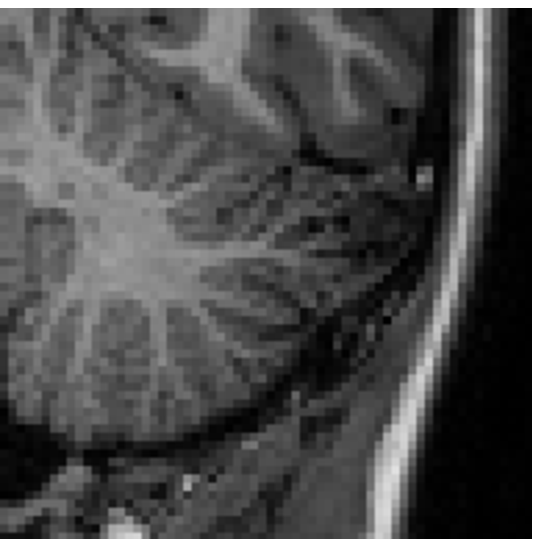}}
  \centerline{(g) Ground Truth}\medskip
\end{minipage}
\vspace{-0.8em}
\caption{Visual comparison of SR results for one coronal slice of the Kirby 21 MRI dataset  \cite{Landman2011} (SR factor 2).}
\label{fig:kirbyCoronal}
\end{figure*}

In Fig.~\ref{fig:mouseSagittal} regions of interest (ROIs) from the sagittal view of a mouse brain and in Fig.~\ref{fig:kirbyCoronal} ROIs from the coronal view of a human brain from the Kirby 21 dataset are visualized.
The SR methods are successful in learning and recovering high-frequency information lost in the tricubically upsampled volumes (Fig.~\ref{fig:mouseSagittal}b and Fig.~\ref{fig:kirbyCoronal}b). 
The cerebellar sub-structure (arbor vitae) of the mouse brain is considerably sharper in the VSRF ROI (Fig.~\ref{fig:mouseSagittal}i). The other 3-D SR results (Fig.~\ref{fig:mouseSagittal}f,g,h) are blurrier than VSRF with less distinct contours, e.g. VA+ shows some artifacts around the arbor vitae. 
Some fine structures of the human cerebellum (Fig.~\ref{fig:kirbyCoronal}g) which are barely visible using tricubic upsampling (Fig.~\ref{fig:kirbyCoronal}b) regain sharpness to some extent with the SR methods. VSRF (Fig.~\ref{fig:kirbyCoronal}f) generates a more precise visualization compared to the other 3-D methods due to artifacts of the VA+ and the fuzziness of the VANR and NLMU results.
VSRF seems to better utilize the features to learn the characteristics of the MRI dataset than VANR and VA+. Further, the choice of the median ensemble model added additional stability against outliers.
The pseudo 3-D methods are also blurrier than VSRF, especially the SRCNN that over-emphasizes the bright area around the arbor vitae. We observe an increase in image quality from the 2-D SRF (Fig.~\ref{fig:mouseSagittal}c) to the pseudo 3-D SRF to VSRF. 
We partly attribute the improvement of VSRF in all three directions compared to SRF and the pseudo 3-D methods to the use of a cubic neighborhood as it better exploits the spacial information contained in the volumes, while the 2-D SRF only captures in-plane relationships. The pseudo 3-D methods average the results of the 2-D methods of the three views, hence recover more information than the 2-D methods but are blurrier than the 3-D methods. 

\begin{figure}[t]
	\scriptsize
	\centering
	\mbox{
	\hspace{-1.3em}
	\subfloat{   	
    		\begin{tikzpicture}
			\begin{axis} [scale = 0.43, 	
			xmajorgrids,
			ymajorgrids,
			xmax   = 13,
			xmin   = 1,
			ylabel = PSNR (dB),
			xlabel = Number of training volumes,
			xtick = {1,3,5,7,9,11,13},
			ytick = {35,36,37,38,39,40},
			ymax   = 40,
			legend style={font=\tiny, style={at={(0.955,0.425)}}, row sep=-0.12cm, inner sep=0pt, line width=0.5pt,}, 
			]
				\addplot coordinates {
				(1,38.74)
				(3,39.08)
				(5,39.26)
				(7,39.35)
				(9,39.42)
				(11,39.42)
				(13,39.46)
					};
				\addplot[mark=none, red, line width=0.8pt] coordinates {
				(1,36.94)
				(13,36.94)};
				\addplot[mark=none, ggreen, line width=0.8pt] coordinates {
				(1,34.94)
				(13,34.94)};
				\legend{VSRF,NLMU,Tricubic}
			\end{axis}
	\end{tikzpicture}
	}\hspace{-0.8em}
	\subfloat{
    		\begin{tikzpicture}
			\begin{axis} [scale = 0.43, 
			xmajorgrids,
			ymajorgrids,
			xmax   = 13,
			xmin   = 1,
			ylabel = SSIM,
			xlabel = Number of training volumes,
			xtick = {1,3,5,7,9,11,13},
			ytick = {0.965, 0.97, 0.975, 0.98},
			yticklabel style={/pgf/number format/precision=4},
			ymax   = 0.9825,
			legend style={font=\tiny, style={at={(0.955,0.425)}}, row sep=-0.12cm, inner sep=0pt, line width=0.5pt,}, 		
			]
				\addplot coordinates {
				(1,0.9779)
				(3,0.9792)
				(5,0.9799)
				(7,0.9801)
				(9,0.9803)
				(11,0.9803)
				(13,0.9804)
					};
				\addplot[mark=none, red, line width=0.8pt] coordinates {
				(1,0.9721)
				(13,0.9721)};
				\addplot[mark=none, ggreen, line width=0.8pt] coordinates {
				(1,0.9637)
				(13,0.9637)};
				\legend{VSRF,NLMU,Tricubic}
					\end{axis}
				\end{tikzpicture}
	}\hspace{-1.3em}
	}
	\vspace{-0.8em}
	\caption{Influence of the number of training volumes for VSRF on the mouse brain dataset (SR factor 2) evaluated with mean PSNR (left) and SSIM (right). NLMU \cite{Manjon2010} and tricubic upsampling results are plotted for comparison.}
	\label{fig:variNtrainVol}
\end{figure}
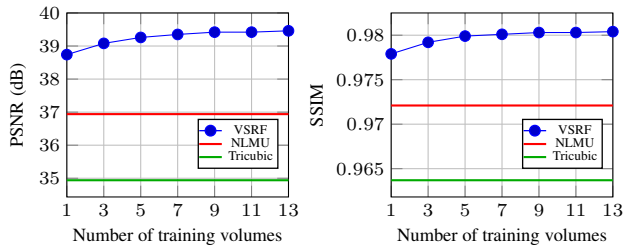
\setlength{\textfloatsep}{5pt plus 1.0pt minus 2.0pt}
\begin{figure}[t]
	\scriptsize
	\centering
	\mbox{
	\hspace{-1.3em}
	\subfloat{ 
\begin{tikzpicture}
    \begin{axis}[
    	width  = 0.25*\textwidth,
        height = 4cm,
        major x tick style = transparent,
        ybar,
        bar width=0.6cm,
    	x=1.5cm, 
    	enlarge x limits={abs=0.7cm}, 
    	enlarge y limits=false,
        ymajorgrids = true,
        ylabel = {PSNR (dB)},
      	ymin=25,
        ymax=48,
        symbolic x coords= {Dev, DevEdge}, 
        xtick = data,
        nodes near coords,
        every node near coord/.append style={font=\tiny},
        nodes near coords align={vertical},
        legend columns=-1, 
        legend style={font=\tiny}, 
        legend pos=north west
        ]
        \addplot [style={bblue,fill=bblue,mark=none}] table[x=Features, y=PSNR_Avg] {data_Features_Ensemble_VSRF.dat};
                \addplot [style={rred,fill=rred,mark=none}] table[x=Features, y=PSNR_Med] {data_Features_Ensemble_VSRF.dat};
        \legend{Average,Median}
    \end{axis}
\end{tikzpicture}
}\hspace{0.8em}
\subfloat{   
	\begin{tikzpicture}
    \begin{axis}[
    	width  = 0.25*\textwidth,
        height = 4cm,
        major x tick style = transparent,
        ybar,
        bar width=0.6cm,
    	x=1.5cm, 
    	enlarge x limits={abs=0.7cm}, 
    	enlarge y limits=false,
        ymajorgrids = true,
        ylabel = {SSIM},
        ymin=0.975,
        ymax=0.983,
        yticklabel style={/pgf/number format/precision=4},
        symbolic x coords= {Dev, DevEdge},  
        xtick = data,
        nodes near coords={\pgfmathprintnumber[fixed zerofill, precision=4]{\pgfplotspointmeta}},
        every node near coord/.append style={font=\tiny},
        nodes near coords align={vertical},
        legend columns=-1,
        legend style={font=\tiny}, 
        legend pos=north west
        ]
        \addplot [style={bblue,fill=bblue,mark=none}] table[x=Features, y=SSIM_Avg] {data_Features_Ensemble_VSRF.dat};
                \addplot [style={rred,fill=rred,mark=none}] table[x=Features, y=SSIM_Med] {data_Features_Ensemble_VSRF.dat};
        \legend{Average,Median}
    \end{axis}
\end{tikzpicture}
	}\hspace{-1.3em}
	}
	\vspace{-0.8em}
	\caption{Influence of the ensemble model (average and median) and features (Dev and DevEdge) for VSRF on the mouse brain dataset (SR factor 2) evaluated with mean PSNR (left) and SSIM (right). }
	\label{fig:variFeaturesVRSF}
\end{figure}
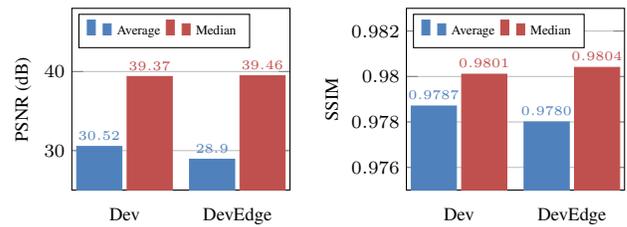

\textbf{Influence of Parameters for VSRF.}
We evaluated the influence on the image quality by the number of training volumes for VSRF, see Fig.~\ref{fig:variNtrainVol}. 
We observed a rapid increase in performance for up to five volumes. More volumes led to minor improvements reaching the highest PSNR and SSIM values for 13 volumes.
Even by using only one training volume, VSRF outperformed tricubic upsampling and NLMU, which do not require training data. Hence, VSRF indicated to be relatively robust with a small amount of training data.
Fig.~\ref{fig:variFeaturesVRSF} showed the clear advantage of the median ensemble for VSRF compared to the average ensemble regarding PSNR and SSIM,
partly due to patch artifacts caused by the average ensemble. 
The feature set DevEdge yielded additional improvement compared to Dev for the median ensemble.

\textbf{Computation Time.} 
The VSRF is very fast in training and inference. 
The inference for the mouse brain MRI dataset required less than $1 \,\mathrm{min}$ per volume and the training of the 13 volumes took about $1 \,\mathrm{h}$ (or about $20 \,\mathrm{min}$ in parallel) with an Intel i7 CPU 3.4 GHz. VANR and VA+ with about $20 \,\mathrm{s}$ per volume were even faster in inference but noticeable slower in training due to the dictionary learning (VANR required $5\,\mathrm{h}\, 15 \,\mathrm{min}$ and VA+ $7\,\mathrm{h}\, 32 \,\mathrm{min}$). The execution time for NLMU took about $90 \,\mathrm{s}$ per volume.
SRCNN required around 3 days for training on a GPU GeForce GTX 1080 and pseudo 3-D SRCNN about $10 \,\mathrm{min}$ per volume for inference using the MATLAB reference implementation of \cite{Dong2014}. 
\section{Conclusion}
\label{sec:conclusion}
We presented a volumetric SR method for brain MRI based on random forests that learn mappings of 3-D LR to HR patches. In the experiments with the mouse brain and the Kirby 21 human brain MRI datasets, our VSRF approach demonstrated visually and quantitatively an improvement of the image quality compared to the state-of-the-art and achieved fast training and inference performance with a small amount of training data. 
The proposed approach for MRI resolution enhancement may be utilized to remarkably reduce MRI acquisition time with smaller loss of image quality, which makes adaption into clinical workflows appealing. 
%
%
\bibliographystyle{IEEEbib}
\bibliography{refs}

\end{document}